\documentclass[twoside,11pt]{article}

\usepackage{jmlr2e}

\usepackage{longtable}
\usepackage{caption} 
\ShortHeadings{Attack-Centric Approach for Evaluating Transferability of Adversarial Samples}{Idika and Akturk}
\firstpageno{1}

\begin{document}

\title{Attack-Centric Approach for Evaluating Transferability of Adversarial Samples in Machine Learning Models}

\author{\name Tochukwu Idika \email Tochukwu.Idika@veteransunited.com \\
       \addr Veterans United\\
       Columbia, MO 65201, USA
       \AND
       \name Ismail Akturk \email ismail.akturk@ozyegin.edu.tr \\
       \addr Computer Science Department\\
       Ozyegin University\\
       Cekmekoy, Istanbul 34794, Turkey}

\editor{}

\maketitle

\begin{abstract}
Transferability of adversarial samples became a serious concern due to their impact on the reliability of machine learning system deployments, as they find their way into many critical applications. Knowing factors that influence transferability of adversarial samples can assist experts to make informed decisions on how to build robust and reliable machine learning systems. The goal of this study is to provide insights on the mechanisms behind the transferability of adversarial samples through an attack-centric approach. This attack-centric perspective interprets how adversarial samples would transfer by assessing the impact of machine learning attacks (that generated them) on a given input dataset. To achieve this goal, we generated adversarial samples using attacker models 
and transferred these samples to victim models. 
We analyzed the behavior of adversarial samples on victim models and outlined four factors that can influence the transferability of adversarial samples. Although these factors are not necessarily exhaustive, they provide useful insights to researchers and practitioners of machine learning systems.  

\end{abstract}


\section{Introduction}

\begin{figure}[h]
	\centering
	\includegraphics[width=0.8\columnwidth]{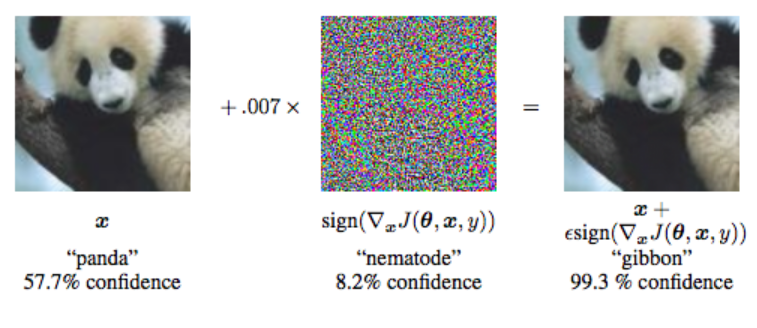}
	\caption{By adding an unnoticeable perturbation to an image of "panda", an adversarial sample is created, and it was misclassified as "gibbon" by the trained network. (Image credit: ~\cite{goodfellow2015})\label{fig:adversarial-input}}
\end{figure}

Machine Learning (ML) applications recently demonstrated widespread adoption in many critical missions, as a way to deal with large-scale and noisy datasets efficiently, in which human expertise cannot be used due to practical reasons. Although ML-based approaches have achieved impressive results in many data processing tasks, including classification, and object recognition, they have been shown to be vulnerable to small adversarial perturbations, and thus tend to misclassify, or not able to recognize minimally perturbed inputs. Figure~\ref{fig:adversarial-input} illustrates how an adversarial sample can be generated by adding a small perturbation, and as a result can get misclassified by a trained Neural Network (NN). 

Adversarial perturbation can be achieved  either through \emph{white-box} or \emph{black-box} attacks. In the threat model of \emph{white-box} attacks, an attacker is supposed to have full knowledge of the target NN model, including the model architecture and all relevant hyperparameters. For the \emph{black-box} attacks, an attacker has no access to the NN model and associated parameters; thus, an attacker relies on generating adversarial samples using the NN model on hand (known as \emph{attacker model}), and then uses these adversarial samples on the target NN model (known as \emph{victim model}). White-box attacks are considered to be difficult to launch in real world scenarios, as it is often not possible for an attacker to have access to full information of the victim model. Thus, in this paper, we focus on \emph{black-box} attacks which are posing practical threats for many ML applications, and evaluate the strategies of generating adversarial samples (which can be used for launching black-box attacks) and their transferability to victim models.

{\bf\textit{Transferability}} is  an ability of an adversarial sample that is generated by a machine learning attack on a particular machine learning model (i.e., on an attacker  model) to be  effective against a different, and potentially unknown machine learning model (i.e., on a victim model). Attacker model refers to the model used in generating the adversarial samples (i.e., malicious inputs that are  modified to yield erroneous output while appearing as unmodified to the human or an agent), whereas the target model refers to the NN model to which the adversarial samples will be transferred. There is a long literature on transferability of adversarial samples and machine learning attacks that generate them; however, they often analyze the transferability from the perspective of a specific network model~\citep{szegedy2014, goodfellow2015, papernot2016, demontis2019}. That is, they have tried to present an explanation on why transferability is able to occur based on the NN model properties (of a given specific target model). Hence, we say that  most research have taken a \emph{model-centric} approach. In contrast, we are presenting an {\bf \textit{attack-centric}} approach, in this paper. In \textit{attack-centric} approach, we provide insights on why adversarial samples can actually transfer by analyzing the adversarial samples generated using different machine learning attacks. A particular insight that we want to build is to see if machine learning attacks and input set have any inherent feature that causes or increases the likelihood of adversarial samples to transfer effectively to the victim models.

In the following, we provide motivation on studying transferability of adversarial samples and exemplify ML-based applications in which they may pose significant security and reliability threats.

\subsection{Motivation for Research on Transferability of Adversarial Samples}

Machine learning has become a driving force for many data intensive innovative technologies in different domains, including (but not limited to) health care, automotive, finance, security, and predictive analytics, thanks to the widespread availability of data sources and computational power allowing to process them in a reasonable time. However, machine learning systems may have security concerns which can be detrimental (and even life threatening) for many application use cases. For motivating the readers regarding the importance of transferability of adversarial samples, and demonstrate the feasibility and possible consequences of machine learning attacks, here we highlight some practical security threats which exploit the transferability of adversarial samples.

\cite{thys2019} generated adversarial samples that were able to successfully hide a person from a person detector camera which relies on a machine learning model. They showed that this kind of attack is feasible to maliciously circumvent surveillance systems and intruders can sneak around undetected by holding on to the adversarial sample/patch in the form of cardboard in front of their body which is aimed towards the surveillance camera. 

Another sector that heavily relies on ML approaches health care due to high volume of data being processed is health care. A particular example of exploiting adversarial samples in this domain is as follows. Dermatologists usually operate under a "fee-for-service" revenue model in which physicians get paid for procedures they perform for a patient. This has caused unethical dermatologists to apply unnecessary procedures to increase their revenue. To avoid frauds in this nature, insurance companies often rely on machine learning models that analyze patient data (e.g., dermatoscopy images) to confirm that suggested procedures are indeed necessary. According to the hypothetical scenario presented by ~\cite{finlayson2018}, an attacker could generate adversarial samples composed of  dermatoscopy images such that when they are analyzed with the machine learning model used by insurance company (victim's model), it would (incorrectly) report that a suggested procedure is appropriate and necessary for the patient.

For security applications that rely on audio commands (which are processed by a ML-based speech recognition system), an attacker can construct adversarial audio samples to be used in breaking into the targeted system. Such an attack, if successful, may lead to information leakage, cause denial of service, or executing unauthorized commands. A feasibility of an attack on speech recognition system is demonstrated by ~\cite{carlini2016} that generated adversarial audio samples (called obfuscated commands) which were used in attacking Google Now's speech recognition system. 

~\cite{jia2017} used Stanford Question Answering Dataset (SQuAD) to test whether text recognition systems can answer questions about paragraphs that contain adversarial sentences inserted by a malicious user. These adversarial samples were automatically generated to mislead the system without changing the correct answers or misleading humans. Their results showed that the accuracy of sixteen published models drops from an average of 75\% F1 score to 36\%, and when the attacker was allowed to add ungrammatical sequences of words, the average accuracy on four of the tested models dropped further down to 7\%. 

As machine learning approaches find their ways into many application domains, the concerns associated with the reliability and security of systems are getting profound. While covering all application areas is out of scope for this paper, our goal is to motivate the study of transferability of adversarial samples to better understand the mechanisms and factors that influence their effectiveness. Without loss of generality, we focus primarily on image classification as a use case to demonstrate the impact of machine learning attacks and their role on effectiveness of transferability of adversarial samples in this paper (though the findings and insights obtained can be generalized for other use cases).


\section{Related Work}
The study of machine learning attacks and transferability of adversarial samples have gained a momentum, following the widespread use of Deep Neural Networks (DNNs) in many application domains. In the following, we detail the recent studies in this area, and discuss their relevance to our work.

\cite{szegedy2014} studied the transferability of adversarial samples on different models that were trained using MNIST dataset. They focused on examining why DNNs were so vulnerable to images with little perturbation. In particular, they examined non-linearity and overfitting in neural networks as the cause of DNNs vulnerability to adversarial samples. Their experiments and methodology, however, were limited to the  NN model characteristics to gain intuition on transferability. 

\cite{goodfellow2015} carried out a new study on transferability of adversarial samples which was built on the previous study of~\cite{szegedy2014}. In contrast, they argued that non-linearity of NN models actually helps to reduce the vulnerability to adversarial samples, and linearity of a model is the cause that makes adversarial samples work. Also, they further suggest that transferability is more likely when the adversarial perturbation or noise is highly aligned with the weight vector of the model. The entire analysis was based on attack called Fast Gradient Sign Method (FGSM) that computes the gradient of the loss function once, and then finds the minimum step size that generates the adversarial samples. 

Another study on transferability was conducted by~\cite{papernot2016} in which they aimed at experimenting how transferability works across traditional machine learning classifiers, such as Support Vector Machines (SVMs), Decision Trees (DT), K-nearest neighbors (KNN), Logistic Regression (LR) and DNNs. Their motivation is to determine if adversarial samples constitute a threat for a specific type or implementation of machine learning model. In other words, they would like to analyze if adversarial samples would be transferred to any of these models; and if so, which of the classifiers (or models) are more prone to such black-box attacks. They also examined intra-technique and cross-technique transferability across the models, and provided in depth explanation on why deep neural network and LR were more prone to intra-technique transferability when compared to SVM, DT, KNN, and LR. However, similar to previous studies, their analysis did not consider the possible impacts of intrinsic properties of attacks on transferability of adversarial samples.

\cite{papernot2017} extended their earlier findings by demonstrating how a black-box attack can be launched on hosting DNN without prior knowledge of the model structure nor its training dataset. The attack strategy employed consists of training a local model (i.e., substitute/attacker model) using synthetically generated data by the adversary that was labeled by the targeted DNN. They demonstrated the feasibility of this strategy to launch black-box attacks on machine learning services hosted by Amazon, Google and MetaMind. Similar study was conducted by~\cite{liu2017}, in which they assumed the model and training process, including both training and test datasets are unknown to them before launching the attack. 

\cite{demontis2019} presented a comprehensive analysis on transferability for both test-time evasion and training-time poisoning attacks. They showed that there are two main factors contributing to the success of the attack that include intrinsic adversarial vulnerability of the target model, and the complexity of the substitute model used to optimize the attack. They further defined three metrics/factors that impacts transferability, which are: i) size of the input gradient, ii) alignment of the input gradients of the loss function computed using the target and the substitute (attacker) models, and iii) variability of the loss landscape. 

All these findings and factors, while essential, are restricted to explain the transferability from the model-centric perspective. However, our investigation is not limited to the assessment of models, but expands the analysis on various attack implementations and the adversarial samples generated to see if there are underlying characteristics that contribute to increasing or decreasing the chances of transferability among NN models.
\section{Machine Learning Attacks}

The adversarial perturbations crafted to generate adversarial samples for fooling a trained network are referred as machine learning attacks. The full list of machine learning attacks presented in the literature is exhaustive, however, we present the subset of attacks analyzed in this work with a brief description of their characteristics in Table~\ref{tab:attacks}.
Following the categorization presented by~\cite{rauber2018}, we categorize the attacks used in this paper into two main families: i) gradient-based, and ii) decision-based attacks. Gradient-based attacks try to generate adversarial samples by finding the minimum perturbation through a gradient descent mechanism. Decision-based attacks involve the use of image processing techniques to generate adversarial samples. It is called decision-based because the algorithms rely on comparing the generated adversarial samples with the original output until misclassification occurs.

\begin{longtable}{| p{.25\textwidth} | p{.18\textwidth} | p{.46\textwidth}|} 

			\hline
			Name of Attack & Attack Family & Short Description\\
			\hline\hline
			Deep Fool Attack & gradient-based &	It obtains minimum perturbation by approximating the model classifier with a linear classifier~\citep{moosavi2016}.\vspace{0.1cm}  \\
			\hline
			Additive Noise Attack &	decision-based & Adds Gaussian or uniform noise and gradually increases the standard deviation until misprediction occurs~\citep{rauber2018}.\vspace{0.1cm}  \\
			\hline
			Basic Iterative Attack & gradient-based & Applies a gradient with small step size and clips pixel values of intermediate results to ensure that they are in the neighborhood of the original image~\citep{kurakin2017}. \vspace{0.1cm} \\
			\hline
			Blended Noise Attack & decision-based & Blends the input image with a uniform noise until the image is misclassified.\vspace{0.1cm}\\
			\hline
			Blur Attack & decision-based & Finds the minimum  blur needed to turn an input image into an adversarial sample by linearly increasing the standard deviation of a Gaussian filter. \vspace{0.1cm}\\
			\hline
			Carlini Wagner Attack &	gradient-based & Generates adversarial sample by finding the smallest noise added to an image that will change the classification of the image~\citep{carlini2017}.\vspace{0.1cm}\\
			\hline
			Contrast Reduction Attack & decision-based & Reduces the contrast of an input image by performing a line-search internally to find minimal adversarial perturbation. \vspace{0.1cm}\\
			\hline
			Search Contrast Reduction Attack& decision-based & Reduces the contrast of an input image by performing a binary search internally to find minimal adversarial perturbation. \vspace{0.1cm}\\
			\hline
			Decoupled Direction and Norm (DDN) Attack & gradient-based & Induces misclassifications with low L2-norm, through decoupling the direction and norm of the adversarial perturbation that is added to the image~\citep{rony2019}. The attack compensates for the slowness of Carlini Wagner attack.\vspace{0.1cm}\\
			\hline
			Fast Gradient Sign Attack & gradient-based & Uses a one-step method that computes the gradient of the loss function with respect to the image once and then tries to find the minimum step size that will generate an adversarial sample~\citep{goodfellow2015}.\\
			\hline
			Inversion Attack &	decision-based & Creates a negative image (i.e., image complement of the original image, in which the light pixels appear dark, and vice versa) by inverting the pixel values~\citep{hosseini2017}.\vspace{0.1cm}\\
			\hline
			Newton Fool Attack & gradient-based & Finds small adversarial perturbation on an input image by significantly reducing the confidence probability~\citep{jang2017}.\vspace{0.1cm}\\
			\hline
			Projected Gradient Descent Attack & gradient-based & Attempts to find the perturbation that maximizes the loss of a model (using gradient descent)  on an input. It is ensured that the size of the perturbation is kept smaller than specified error by relying on clipping the samples generated~\citep{madry2017}.\vspace{0.1cm}\\
			\hline
			Salt and Pepper Noise Attack & decision-based &	Involves adding salt and pepper noise to an image in each iteration until the image is misclassified, while keeping the perturbation size within the specified epsilon $\epsilon$.\vspace{0.1cm}\\
			\hline
			Virtual Adversarial Attack & gradient-based & Calculates untargeted adversarial perturbation by performing an approximated second order optimization step on the Kullback–Leibler divergence between the unperturbed predictions and the predictions for the adversarial perturbation~\citep{miyato2015}. \vspace{0.1cm}\\
			\hline
			Sparse Descent Attack & gradient-based & A version of basic iterative method that minimizes the L1 distance. \vspace{0.1cm}\\ 
			\hline
			Spatial Attack & decision-based & Relies on spatially chosen rotations, translations, scaling~\citep{engstrom2019}.\vspace{0.1cm}\\ 
			\hline \hline

\caption{The machine learning attacks used in this work.}
\label{tab:attacks}
\end{longtable}


\section{Methodology}

In the following, we detail the Convolutional Neural Network (CNN) models, infrastructure and tools used in the evaluation, as well as the procedure employed in carrying out the experiments. 

\subsection{Infrastructure and Tools}

To build, train and test the CNNs that use in our evaluation, we rely on PyTorch and TorchVision. We also use Foolbox~\citep{rauber2018} which is a Python library to generate adversarial samples. It provides reference implementations for many of the published adversarial attacks, all of which perform internal hyperparameter tuning to find the minimum adversarial perturbation. We use Python version 3.7.3 on Jupyter Notebook. We run our experiments on Google Colab which provides an interactive environment that allows to write and execute Python code. It is similar to Jupyter notebook, but rather than being installed locally, it is hosted on the cloud. It is heavily customized for data science workloads, as it contains most of the core libraries used in data science/machine learning research. We used this environment in training the neural network as it provides large memory capacity and access to GPUs, thereby reducing the training time. 

\subsection{CNNs Used in This Study}

Here, we provide the brief description and details of CNNs used in this work. Note that a particular CNN may be in one of two roles, namely it can be either an attacker model (on which the adversarial samples are generated), or a victim model (to which the adversarial samples will be used to attack).

{\bf LeNet:}
It is simple, yet popular CNN architecture that was first introduced in 1995 but came to limelight in 1998 after it demonstrated success in handwritten digit recognition task~\citep{lecun1998}. The LeNet architecture used for this work is slightly modified to train for CIFAR-10 dataset (instead of MNIST).

{\bf AlexNet:}
It is an advanced form of LeNet architecture, with a depth of 8 layers. It showed groundbreaking results in 2012- ILSVRC competition by achieving an error  rate from 25.8\% to 16.4\% on ImageNet dataset with about 60 million trainable parameters~\citep{krizhevsky2017}. It also has different optimization techniques such as dropout, activation function and Local Response (LR) normalization. Since LR normalization had shown minimal (if any) contribution in practice it was not included in the AlexNet model trained for this project.  Aside from the increase in the depth of the  network, another difference between the LeNet and AlexNet model trained in this work is that AlexNet has a dropout layer added to it.

{\bf Vgg-11:}
It was introduced to improve the image classification accuracy on ImageNet dataset by~\cite{simonyan2015}. Compared to LeNet and AlexNet, Vgg-11 has an increased network depth, and  it made use of small (3 x 3) convolutional filters. The architecture secured a second place at the ILVRSC 2014 competition after reducing the error rate on the ImageNet dataset down to 7.3\%. Hence, the architecture is an improvement over AlexNet. There are different variants of Vgg : Vgg-11, 13, 16 and 19. Only Vgg-11 is used in this paper. In addition to being deeper than  AlexNet architecture, batch normalization is also introduced in the Vgg-11 used in this project.

Table~\ref{tab:cnn-models} summarizes the major features of these three CNN models. We choose these models to evaluate how machine learning attacks and corresponding adversarial samples generated respond to these models.

\begin{longtable}{| p{.08\textwidth} | p{.072\textwidth} | p{.12\textwidth}| p{.109\textwidth} | p{.125\textwidth} | p{.065\textwidth} | p{.12\textwidth} | p{.1\textwidth}|} 
	\hline
CNN& \# Conv. Layers&\# Inner activation func., type&Output activation func.& \# Pooling Layers, type& \# FC Layers&\# Dropout Layers (\%)&\# BatchNorm Layers \vspace{0.1cm}\\
\hline
LeNet&2&4, RELU &Softmax& 2, maxpool& 3 &None & None \vspace{0.1cm}\\
\hline
AlexNet&5&7, RELU&Softmax &3, maxpool& 3 & 2 (\%0.5)& None \vspace{0.1cm}\\
\hline
Vgg-11& 8&8, RELU&Softmax&4, maxpool& 3 & 2 (\%0.5) & 8 \vspace{0.1cm}\\
\hline
\hline
\caption{Features of the CNN models used in this paper.}
\label{tab:cnn-models}
\end{longtable}

\subsection{Data Processing and Training}

{\bf Dataset:} We used CIFAR-10 dataset~\citep{Krizhevsky2009} for our analysis, since it is arguably one of the most widely used dataset in the field of image processing and computer vision research. It contains 60,000 images which belong to one of ten classes. Training dataset contains 45,000 images, validation dataset has 500 images, whereas testing dataset contains 10,000 images. To generate adversarial samples, 500 images are selected from the testing dataset (50 images picked from each class to have a balanced dataset).

\noindent {\bf Preprocessing:} At the very beginning, we performed training transformations, including random rotation, random horizontal flip,  random cropping, converting the dataset to tensor and normalization. Likewise, we performed test transformations, including converting the dataset to tensor, and normalized it. Random rotation and horizontal flip introduce complexity to the input data which helps the model to learn in a more robust way. It is necessary to convert inputs to tensor because PyTorch works with tensor objects. The three channels are normalized (dividing by 255) to increase learning accuracy. Final step of data pre-processing was forming a batch size of 256 and creating a data loader for train and validation data (the method loads 256 images in each iteration during the training and validation). We choose batch size of 256 as it is large enough to make the training faster. 

\noindent {\bf Training:} For the training, we first created the network model which comprises of feature extraction, classification and forward propagation. In each epoch, we calculated the training loss, training accuracy, validation loss and validation accuracy.  To perform training, we specified the following parameters for the train function: model, training iterator, optimizer (Adam optimizer) and criterion (cross entropy loss criterion). To perform validation, we specified the following parameters to the evaluation function: model, validation iterator, and criterion (cross entropy loss criterion). After completing training phase, we saved parameter values for the given model. 

\begin{longtable}{| p{.3\textwidth} | p{.2\textwidth}| p{.2\textwidth} | p{.2\textwidth} |} 
	\hline
	Characteristics & LeNet & AlexNet & Vgg-11 \vspace{0.1cm}\\
	\hline
	\hline
	Epoch number &	25 & 25 & 10 \vspace{0.1cm}\\
	\hline
	Training loss &	0.953 &	0.631 &	0.244 \vspace{0.1cm}\\
	\hline
	Validation loss & 0.956 & 0.695 & 0.468 \vspace{0.1cm}\\
	\hline
	Training accuracy &	66.34\% & 78.34\%&	91.94\% \vspace{0.1cm}\\
	\hline
	Validation accuracy & 66.70\% &	76.74\%&87.11\% \vspace{0.1cm}\\
	\hline
	Testing accuracy &	66.64\% &76.03\%&	85.87\% \vspace{0.1cm}\\	
	\hline
	\hline
	\caption{Training characteristics for NN models.}
	\label{tab:training-characteristics}
\end{longtable}

The final step is the testing stage. To test the trained models, we loaded in the saved model parameters, including trained weights. Then, we checked for testing accuracy of the networks. Table~\ref{tab:training-characteristics} summarizes the training characteristics and reports train, validation and testing accuracy obtained.

\subsection{Adversarial Samples Generation}

{\bf Machine learning attacks:} Table~\ref{tab:attacks} detailed  17 unique machine learning attacks employed in the evaluation. However, for some of the attacks, more than one norms (L0, L1, L-infinity) are used for estimating the error ($\epsilon$), thus increasing the number of unique attacks evaluated to 40. For the sake of brevity, we enumerate the attacks ranging from 1 to 40 (as listed in Table~\ref{tab:attack-enumeration}), and used this enumeration as labels, instead of providing the full name and used-norm when showing the results in the following Figures.

\begin{longtable}{| p{.05\textwidth} | p{.3\textwidth}| p{.055\textwidth} || p{.05\textwidth} | p{.3\textwidth}| p{.055\textwidth} | } 
	\hline
	Label & Attack Name & Norm & Label & Attack Name & Norm \\
	\hline
	\hline
	1& Deep Fool Attack&	L-inf & 21& BSCR Attack& L2\\
	\hline
	2& Deep Fool Attack&	L2 & 22& BSCR Attack& L-inf\\
	\hline
	3& Additive Gaussian Noise (AGN) Attack&	L2 & 23& Linear Search Contrast Reduction (LSCR) Attack&	L1\\
	\hline
	4& Additive Uniform Noise (AUN) Attack&	L2 & 24& LSCR Attack& L2\\
	\hline
	5& AUN Attack&	L-inf & 25& LSCR Attack& L-inf\\
	\hline
	6& Repeated AGN Attack&	L2 & 26& Decoupled Direction and Norm Attack& L2\\
	\hline
	7& Repeated AUN Attack&	L2 & 27& Fast Gradient Sign Attack& L1\\
	\hline
	8& Repeated AUN Attack&	L-inf & 28& Fast Gradient Sign Attack& L2\\
	\hline
	9& Basic Iterative Attack& L1 & 29& Fast Gradient Sign Attack& L-inf\\
	\hline
	10& Basic Iterative Attack&	L2& 30& Inversion Attack& L1\\
	\hline
	11& Basic Iterative Attack&	L-inf& 31& Inversion Attack& L2\\
	\hline
	12& Blended Uniform Noise Attack& L1 & 32& Inversion Attack& L-inf\\
	\hline
	13& Blended Uniform Noise Attack& L2 & 33& Newton Fool Attack& L2\\
	\hline
	14& Blended Uniform Noise Attack& L-inf & 34& Projected Gradient Descent Attack& L1\\
	\hline
	15& Blur Attack& L1 & 35& Projected Gradient Descent Attack& L2\\
	\hline
	16& Blur Attack& L2 & 36& Projected Gradient Descent Attack& L-inf\\
	\hline
	17& Blur Attack& L-inf & 37& Salt and Pepper Attack& L2\\
	\hline
	18& Calini Wagner Attack& L2 & 38& Sparse descent Attack& L1\\
	\hline
	19& Contrast Reduction Attack& L2 & 39& Virtual adversarial Attack&	L2\\
	\hline
	20& Binary Search Contrast Reduction (BSCR) Attack&	L1 & 40& Spatial Attack&	N/A\\
	\hline
	\caption{Labels of attacks and norms used to generate adversarial samples.}
	\label{tab:attack-enumeration}
\end{longtable}

{\bf Adversarial Sample Formulation:} Given a classification function $f(x)$, class $C_x$, adversarial classification function $f(x\prime)$, distance $D(x, x\prime)$  and epsilon $\epsilon$ (smallest allowable perturbation or error), adversarial sample $x$ can be mathematically expressed as:

\[
f(x)\; = \;C_x \land f(x\prime)\;\neq\;C_x \land D(x,x\prime) \leq \epsilon.
\]

To craft adversarial samples via Foolbox~\citep{rauber2018}, we need to specify a criterion that defines the impact of adversarial action (misclassification in our case), and a distance measure that defines the size of a perturbation (i.e., L1-norm, L2-norm, and/or L-inf which must be less than specified $\epsilon$). Then, these are taken into consideration in an attacker model to generate an adversarial sample. 
The following equation shows the general distance formula. Depending on the value of p, L1, L2 or L-inf norm is obtained. 

\[
||x - \hat{x}||_p \; = \; (\; \sum_{i=1}^{d} | x_i = \hat{x_i}|^p \;)^{1/2}
\]

We picked the value of epsilon as 1.0, since it allows to generate a significant number of adversarial samples for all the attack methods used. Because it takes a lot of time to generate adversarial samples using the attack algorithms, we used 500 balanced inputs (i.e., 50 images from each of the 10 classes) from the test data.

\begin{figure}[h]
	\centering
	\includegraphics[width=0.6\columnwidth]{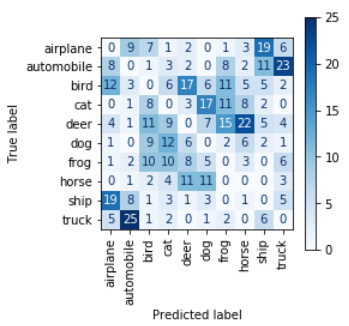}
	\caption{Confusion matrix of adversarial samples generated using Deep Fool attack with L-inf norm on LeNet. 	\label{fig:confusion-linf}}
	\vspace{-0.2cm}
\end{figure}

\begin{figure}[h]
	\centering
	\includegraphics[width=0.6\columnwidth]{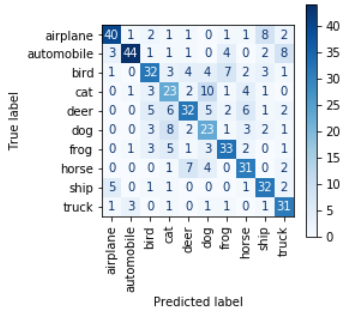}
	\caption{Confusion matrix of adversarial samples generated using Additive Gaussian Noise attack with L2 norm on LeNet.\label{fig:confusion-l2}}
	\vspace{-0.2cm}
	
\end{figure}

To demonstrate how well adversarial samples transfer, we use a confusion matrix as a visual guide. In a given confusion matrix, each row represents instances in a predicted class, whereas each column represents instances in an true/actual class that a given input belongs. The diagonal of the confusion matrix shows the number of each class that were correctly predicted after an attack is launched. For example, Figure~\ref{fig:confusion-linf} shows a confusion matrix of adversarial samples generated by using Deep Fool attack (with L-inf norm) on LeNet. It has all zero entries on the diagonal which means that the inputs (i.e., adversarial samples) were misclassified in all classes. This implies that the attack that generated the adversarial samples is very powerful since they were all misclassified. On the other hand, Figure~\ref{fig:confusion-l2} 
shows a confusion matrix of adversarial samples generated by using Gaussian Noise attack (with L2 norm) on LeNet. In this confusion matrix, however, the diagonal has non-zero, larger positive entries that illustrates the attack used in generating the adversarial samples are less powerful leading as many of the samples correctly classified. 

\subsection{Experimental Procedure}

Here, we describe the procedure in performing the analysis and generating the results shown in the Evaluation. First, the adversarial samples are generated by using the attack model and original dataset on an attacker model (which can be one of the LeNet, AlexNet, or Vgg-11 at any given scenario). Once the adversarial samples are generated on the attacker model, they are used on the victim models (which can be one of the LeNet, AlexNet or Vgg-11). Then, the statistics regarding the number of mispredictions, as well as their prediction classes are collected. We also calculate the Structural Similarity Index Measure (SSIM) between adversarial samples and the original sample to compare how visually similar they are (SSIM value ranges from 0 to 1; the higher value indicates more similarity). This measure has been used in the literature to correlate more with human perception than Mean Absolute Distance (MAD). Hence, it serves as a metric for estimating how much perturbed (adversarial) and the original image will differ visually.


\section{Evaluation}

We obtained three kinds of results using adversarial samples generated on attacker models: i) number of mispredictions when adversarial samples are used on victim models; ii) the classes that (mis)predictions belong when adversarial samples are used on victim models; and iii) SSIM value between original and adversarial samples. 
We used these results to assess the effectiveness of attacks used in generating adversarial samples. This assessment led us to identify four main factors that contribute immensely towards the transferability of adversarial samples. In the following, we discuss these factors and provide results obtained to backup our findings for each factor's implication. 
  

\subsection{Factor 1: The attack itself}
We observed that some of the attacks used in generating adversarial samples are just more powerful than others (regardless of the victim model). That is, the adversarial samples generated by these attacks are easily transferable, hence leading to high number of misprediction on the target model. 

\begin{figure}[h]
	\centering
	\includegraphics[width=1\columnwidth]{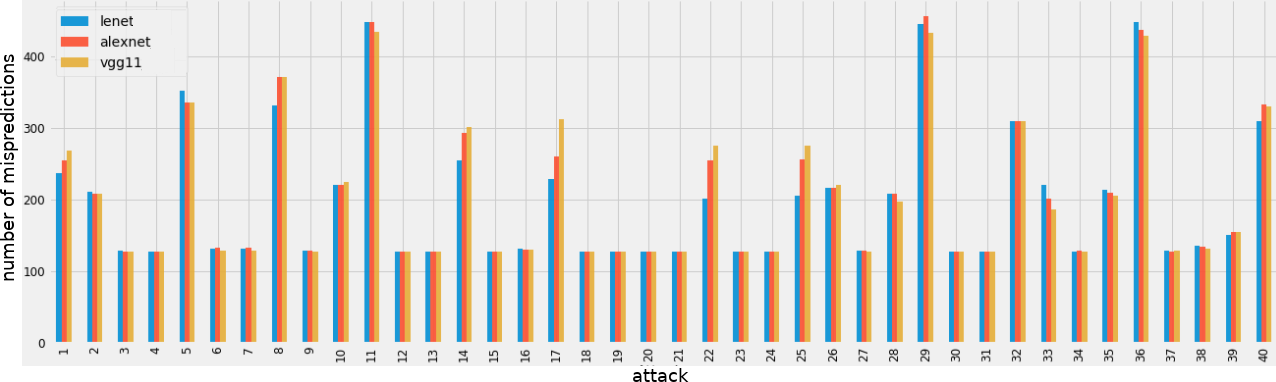}
	\caption{Average number of mispredictions for adversarial samples transferred to the LeNet, AlexNet and Vgg-11. \label{fig:attacks}}
\end{figure}

Figure~\ref{fig:attacks} shows that the attacks with labels 1, 5, 8, 11, 14, 17, 25, 29, 32, 36, and 40 have higher number of  mispredictions when adversarial samples are used on victim models. Hence, those attacks are more powerful. Further, attacks with labels 11, 29 and 36 appear to have the highest number of mispredictions (on any victim model). This result shows that the transferability of an adversarial sample highly depends on the attack that generated the given adversarial sample.

\subsection{Factor 2: Norm Used in the Attack}
\begin{figure}[h]
	\centering
	\includegraphics[width=1\columnwidth]{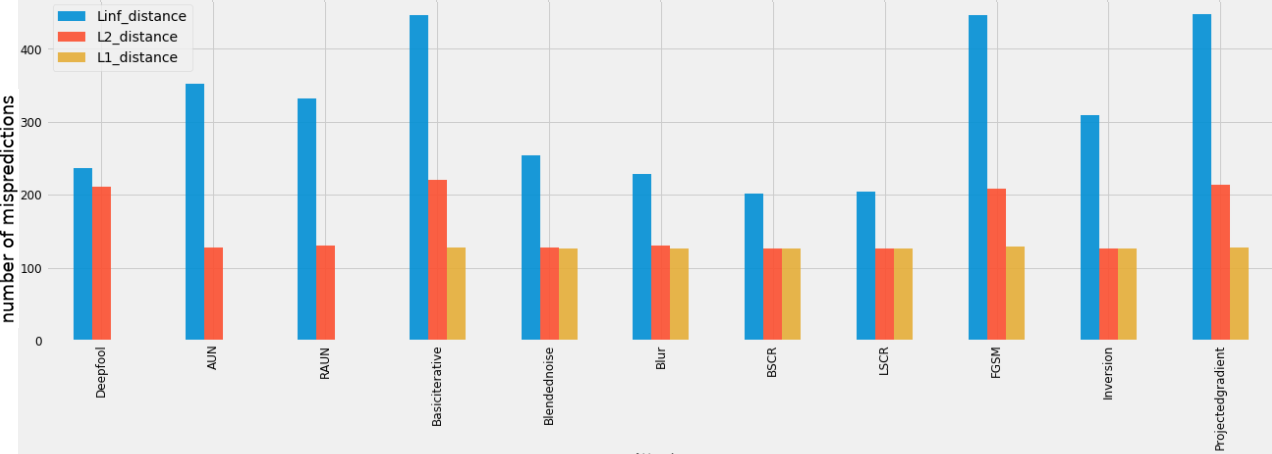}
	\caption{Average number of mispredictions per attack for adversarial samples generated on LeNet. \label{fig:lenet-attacker-distances}}
\end{figure}

\begin{figure}[h]
	\centering
	\includegraphics[width=1\columnwidth]{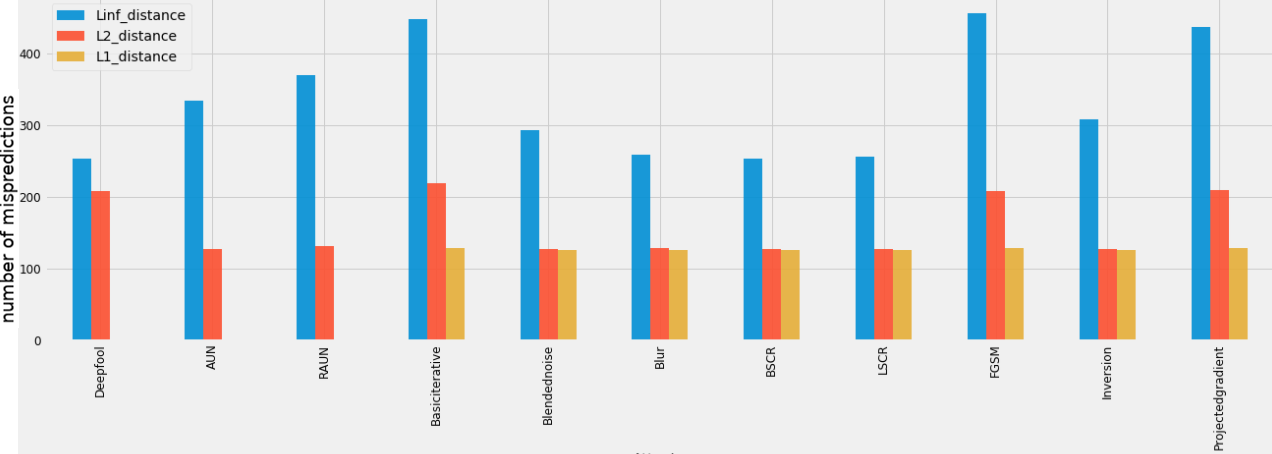}
	\caption{Average number of mispredictions per attack for adversarial samples generated on AlexNet. \label{fig:alexnet-attacker-distances}}
\end{figure}

\begin{figure}[h]
	\centering
	\includegraphics[width=1\columnwidth]{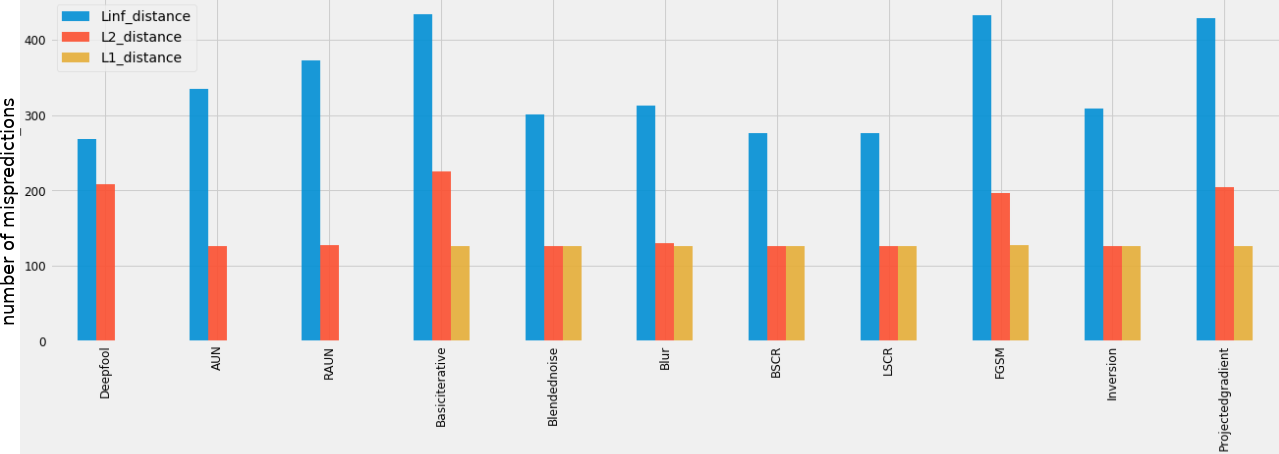}
	\caption{Average number of mispredictions per attack for adversarial samples generated on Vgg-11. \label{fig:vgg11-attacker-distances}}
\end{figure}

We observed that a particular attack that uses different norm to generate adversarial samples yielded varying degree of transferability. In general, the attacks that use L-inf tend to produce adversarial samples that exhibit higher number of mispredictions compared to attacks using L2 and L1. Figures~\ref{fig:lenet-attacker-distances},~\ref{fig:alexnet-attacker-distances} and \ref{fig:vgg11-attacker-distances} show results for attacks that use different norms when generating adversarial samples. In particular, Figure~\ref{fig:lenet-attacker-distances} shows the average number of mispredictions per attack for adversarial samples that are generated on LeNet. Among the attacks, Deep Fool, AUN and RAUN are implemented by using just L-inf and L2, whereas the rest have implementation for L1, L2 and L-inf norms. Clearly, the adversarial samples generated with L-inf norm have stronger ability to transfer, compared to the ones generated with L1 and L2 norms. Likewise, Figure~\ref{fig:alexnet-attacker-distances} and~\ref{fig:vgg11-attacker-distances} show the average number of mispredictions per attack for adversarial samples that are generated on AlexNet and Vgg-11, respectively. The findings are consistent among the victim models, indicating the norm to be used for a given attack has a significant impact on transferability of adversarial samples.

\begin{figure}[h]
	\centering
	\includegraphics[width=1\columnwidth]{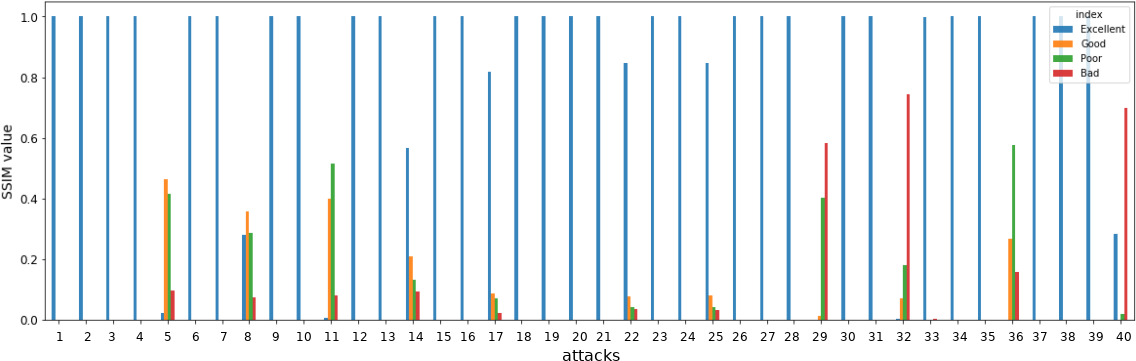}
	\caption{SSIM values for adversarial samples generated on AlexNet.  	\label{fig:ssim}}
\end{figure}

While L-inf norm yields adversarial samples to transfer better compared to other norms, it should be noticed that the disturbance made to an input sample may become more pronounced. Comparing SSIM values of adversarial samples generated by using different norms shows that L-inf always produces significantly perturbed samples. In Figure~\ref{fig:ssim}, the range for SSIM values are labeled as: Excellent = ( 0.75 $\leq$ SSIM $\leq$ 1.0 ), Good = ( 0.55 $\leq$ SSIM $\leq$0.74 ),  Poor = (0.35 $\leq$ SSIM $\leq$ 0.54), and Bad = (0.00 $\leq$ SSIM  $\leq$ 0.34). We observed that many of the adversarial samples generated by L-inf norm have lower SSIM, indicating that perturbations made may be perceived by human. Therefore, checking the SSIM values can be used to guide the effectiveness of a given attack. Although an attack aims to maximize the number of mispredictions, it should be considered as a stronger attack if it can keep SSIM higher while yielding higher number of mispredictions, at the same time. 

\subsection{Factor 3: Closeness of the Target Model to the Attacker Model}

\begin{figure}[h]
	\centering
	\includegraphics[width=1\columnwidth]{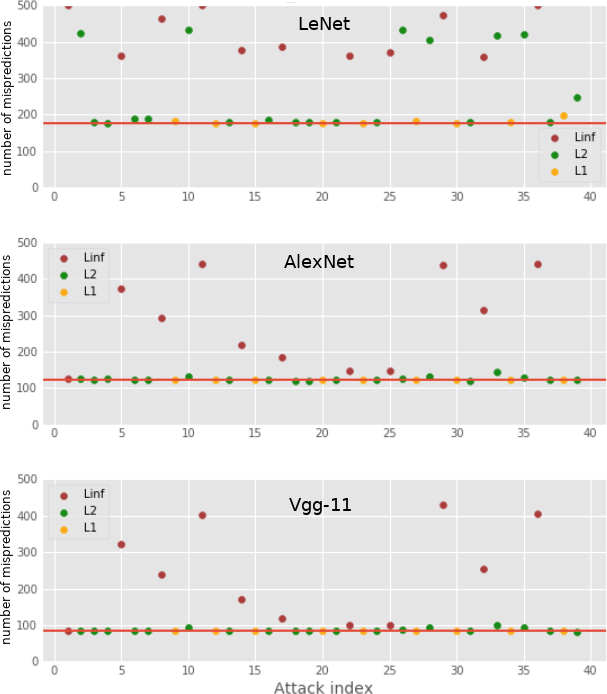}
	\caption{Number of mispredictions for adversarial samples that are generated on LeNet.\label{fig:lenet-attacker-model}}
\end{figure}

\begin{figure}[h]
	\centering
	\includegraphics[width=1\columnwidth]{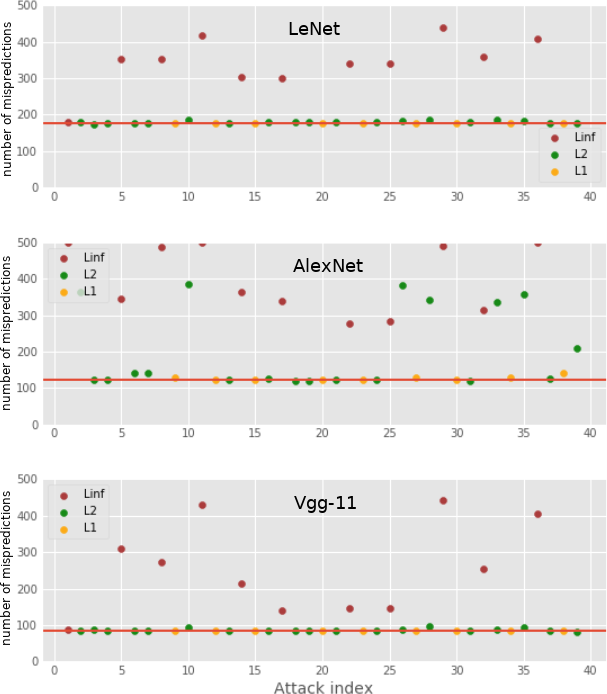}
	\caption{Number of mispredictions for adversarial samples that are generated on AlexNet. \label{fig:alexnet-attacker-model}}
\end{figure}

\begin{figure}[h]
	\centering
	\includegraphics[width=1\columnwidth]{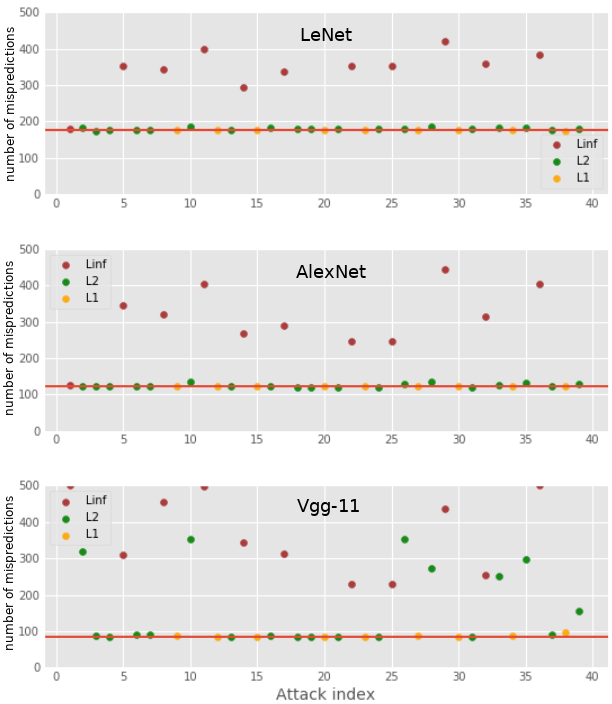}
	\caption{Number of mispredictions for adversarial samples that are generated on Vgg-11.\label{fig:vgg11-attacker-model}}
\end{figure}

Not surprisingly, we observed that adversarial samples yielded higher number of mispredictions for the models on which they were generated (i.e., the case in which attacker and victim models are the same). For example, adversarial samples generated on AlexNet lead to higher number of misprediction when these samples are used on AlexNet, or on a closer model (e.g., a variation of AlexNet). However, when these adversarial samples are used on other (or dissimilar) victim models,  they lead to a comparably lower number of  mispredictions. These findings are shown in Figures~\ref{fig:lenet-attacker-model},~\ref{fig:alexnet-attacker-model} and \ref{fig:vgg11-attacker-model}.

The implication of this factor is that if an attacker can generate adversarial samples on a model that is similar to victim models, then the probability of adversarial samples generated to be transferred effectively is higher. This methodology can be used by industry experts to test how well adversarial samples can transfer  to their ML models. One way to exploit this observation for security-critical applications is to build multiple ML models that are dissimilar in terms of structure, but providing similar prediction accuracy; and then using majority vote (or similar schemes) to decide what should be proper prediction. If a particular attack would transfer and be effective on one of the ML model, (as evident by the analysis) it is very likely that other ML models (which are dissimilar) would be less sensitive to the same attack, providing a way to detect the anomaly and avoid the undesired consequences of adversarial samples. Building ML models that are different in structure, but yielding similar accuracy would be active research direction, not just for security-related concerns, but also be useful for reliability, power management, performance and scalability.

\subsection{Factor 4: Sensitivity of  an Input}
Inherent sensitivity of an input to a particular attack can determine the strength of adversarial sample and how well it can transfer to a victim model. We can summarize our observations about the sensitivity of inputs used in the attacks as follows.

\begin{figure}[h]
	\centering
	\includegraphics[width=1.0\columnwidth]{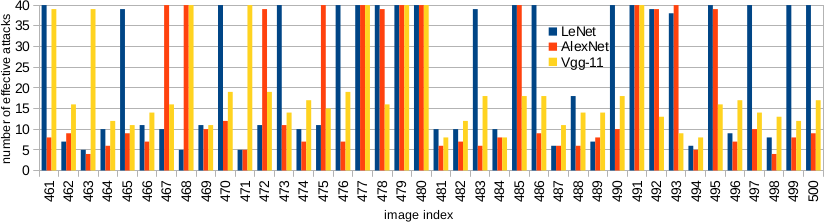}
	\caption{The number of effective attacks (yielding a adversarial sample that would be mispredicted) for a particular input used on Vgg-11 as an attacker model (zoomed in to see last 40 input images). \label{fig:vgg11-misprediction}}
\end{figure}

\begin{enumerate}
	\item Some inputs are very sensitive to almost any attack, thus the adversarial samples generated for them can effectively transfer to victim models (e.g., input images with index 477, 479, 480 and 481 in Figure~\ref{fig:vgg11-misprediction}). 
	\item Some inputs are insensitive to attacks, thus the adversarial samples generated are ineffective and cannot get mispredicted, regardless of the victim model (e.g., input images with index 481, 484, 494 in Figure~\ref{fig:vgg11-misprediction}).
	\item Some inputs are sensitive to specific attacks on a particular victim model, meaning the adversarial samples become effective when they are generated by particular subset of attacks, targeting a particular model (but not effective when used on other models). For example, the input images with index 465 and 467 in Figure~\ref{fig:vgg11-misprediction} become more sensitive (thus corresponding adversarial samples are more effective) when they are transferred to LeNet and AlexNet models, respectively (but not on other models). 
\end{enumerate}

\begin{figure}[h]
	\centering
	\includegraphics[width=1.0\columnwidth]{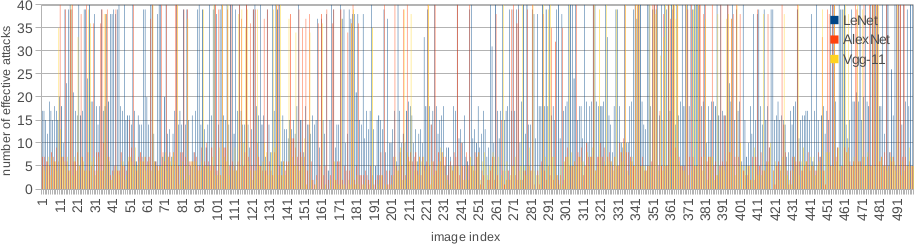}
	\caption{The number of effective attacks (yielding a adversarial sample that would be mispredicted) for a particular input used on LeNet as an attacker model. \label{fig:lenet-misprediction-all}}
\end{figure}
\begin{figure}[h]
	\centering
	\includegraphics[width=1.0\columnwidth]{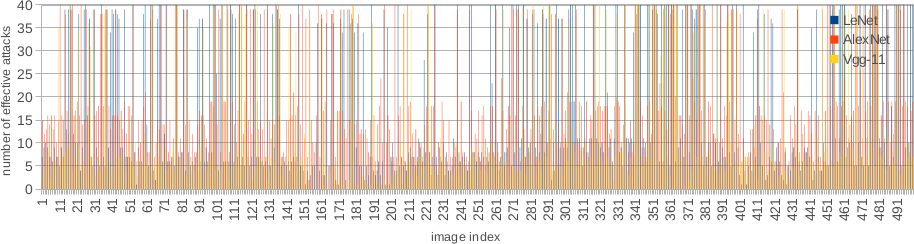}
	\caption{The number of effective attacks (yielding a adversarial sample that would be mispredicted) for a particular input used on AlexNet as an attacker model. \label{fig:alexnet-misprediction-all}}
\end{figure}
\begin{figure}[h]
	\centering
	\includegraphics[width=1.0\columnwidth]{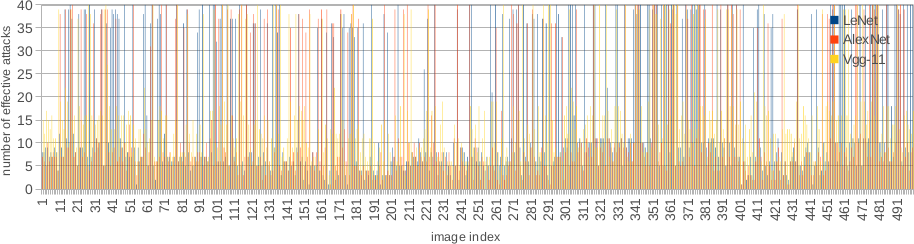}
	\caption{The number of effective attacks (yielding a adversarial sample that would be mispredicted) for a particular input used on Vgg-11 as an attacker model. \label{fig:vgg11-misprediction-all}}
\end{figure}

\begin{figure}[h]
	\centering
	\includegraphics[width=0.7\columnwidth]{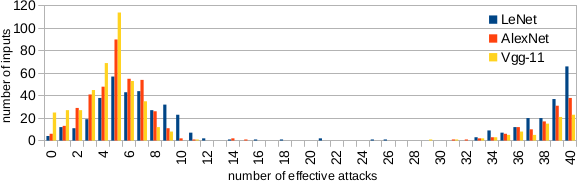}	
	\caption{Histogram that summarizes the sensitivity of inputs to attacks. The x-axis indicates the number of effective attacks for a given input (i.e., generated adversarial sample would transfer to victim model successfully regardless of the attacker model), and y-axis indicates the number of inputs whose adversarial samples (generated by a set of attacks) would transfer effectively to the victim models. \label{fig:collective-histogram} }
\end{figure}

Figure~\ref{fig:vgg11-misprediction} shows the effective number of attacks used to generate adversarial samples on Vgg-11. For better visibility, only the last 40 input images (out of 500) are zoomed in Figure~\ref{fig:vgg11-misprediction} where  the x- axis shows the index of input image and the y-axis shows the number of attacks that lead to misprediction of generated adversarial samples on victim models (please, see Figure~\ref{fig:vgg11-misprediction-all} for all 500 inputs used on Vgg-11). Since there are 40 attacks used to generate adversarial samples, the y-axis can be at most 40 (in which case it would mean that all of the attacks yielded adversarial sample that result in misprediction). The results obtained for the complete 500 input images used are shown in Figure~\ref{fig:alexnet-misprediction-all},~\ref{fig:lenet-misprediction-all} for AlexNet, and LeNet (as attacker model), respectively.

The implication of this factor is that the inherent characteristics of the input may play a role on how effectively the generated adversarial samples would be transferred to victim models. When combined with the strength of an attack, some inputs that are sensitive to the given set of attacks (irrespective of attacker model) may yield more effective adversarial samples than the other inputs. 
Figure~\ref{fig:collective-histogram} illustrates this phenomena. It can be seen that  most of the input images are sensitive to roughly 10 attacks out of the 40 (regardless of the attacker model being used), but relatively very few inputs are very sensitive to all the attacks (23 input images yield adversarial samples that were mispredicted on all the victim models, regardless of attacker model and attack used). 

\section{Conclusion}

In its simplest form, \textit{transferability} can be defined as the ability of adversarial samples generated using the attacker model to be mispredicted when transferred to the victim model. We identified that most of the literature on transferability focuses on interpreting and evaluating transferability from the machine learning model perspective alone, which we refer as model-centric approach. In this work, we took an alternative path that we call attack-centric approach that focuses on investigating machine learning attacks to interpret and evaluate how adversarial samples transfer to the victim models. For each attacker model, we generated adversarial samples that are transferred to the three victim models (i.e., LeNet, AlexNet and Vgg-11). 
We identified four factors that influence how well an adversarial sample would transfer. 
Our hope is that these factors would be useful guidelines for researchers and practitioners in the field to prohibit the adverse impact of black-box attacks and to build more attack resistant/secure machine learning systems. 

\vskip 0.2in
\bibliography{sample}

\end{document}